\documentclass[runningheads]{llncs}

\usepackage{eccv}

\usepackage{hyperref}

\usepackage{orcidlink}

\usepackage{eccvabbrv}
\usepackage{graphicx}
\usepackage{booktabs}
\usepackage[accsupp]{axessibility} 
\usepackage{enumitem}
\usepackage{algorithm}
\usepackage{algorithmic}

\usepackage{booktabs}
\usepackage{graphicx}
\usepackage[table]{xcolor}
\definecolor{hlblue}{RGB}{230, 242, 255}  
\definecolor{hlyellow}{RGB}{255, 249, 230}

\begin{document}

\title{DeforM: Reasoning-Guided Physics-Aware Video Generation via Spatial-Temporal Masking} 

\titlerunning{DeforM}

\author{ Yunyi Li \and Yu Qiao \and Yaohui Wang \and Xinyuan Chen\thanks{Corresponding author.}}

\authorrunning{Y.~Li et al.}

\institute{
Shanghai Artificial Intelligence Laboratory, Shanghai, China\\
Project Page: \url{https://nemosunny.github.io/DeforM-ProjectPage/}
}

\maketitle



\begin{figure}
    \centering
    \includegraphics[width=1.0\linewidth]{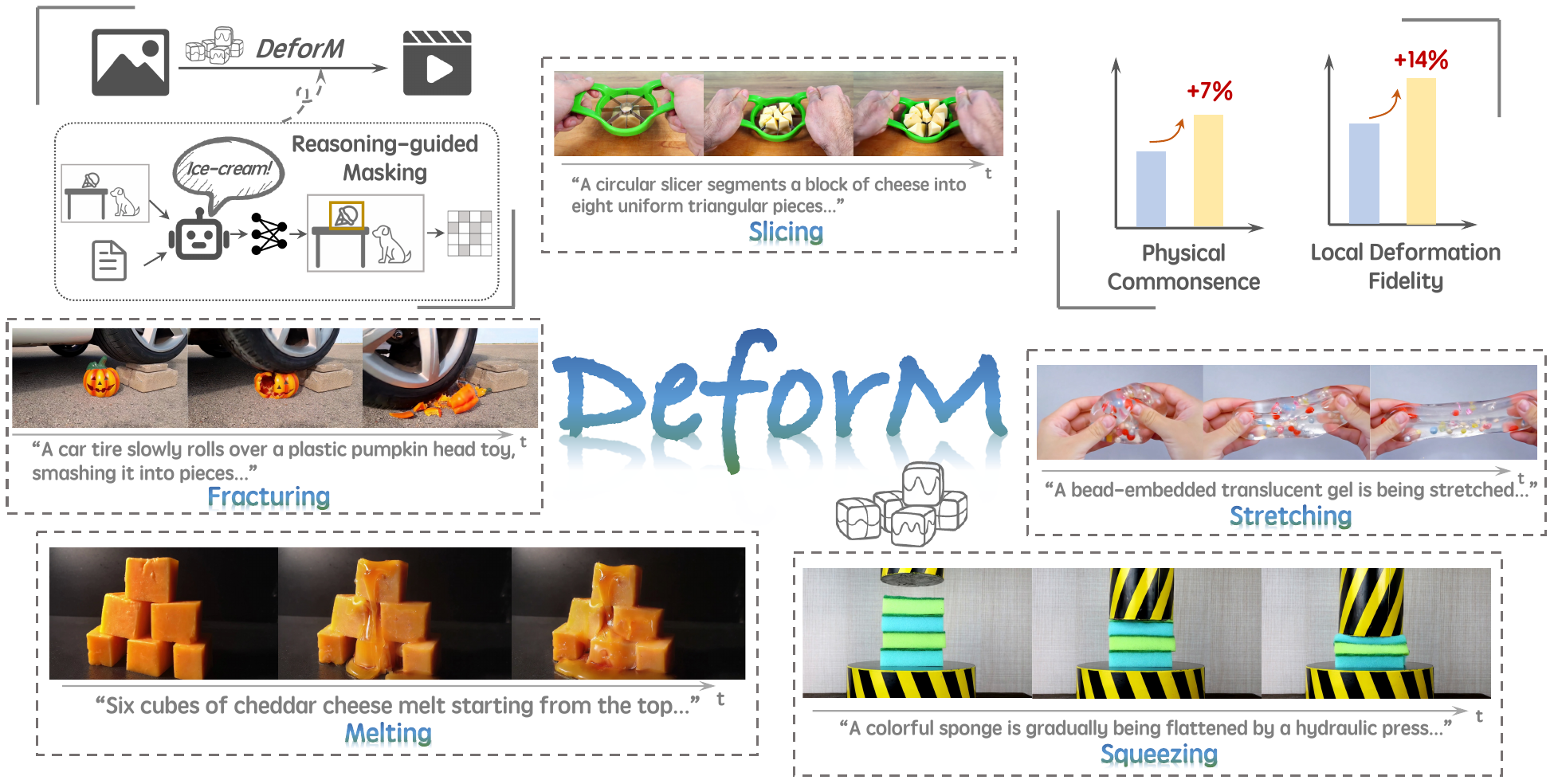}
    \caption{DeforM is an image-to-video physics-aware generation framework. It generates realistic deformation dynamics by leveraging VLM-guided physical reasoning to focus on physics-critical regions.}
    \label{fig:fig1}
    \vspace{-6mm}
\end{figure}

\begin{abstract}
Video generation models achieve high visual quality but often struggle to generate physics-aware videos. Unlike rigid-body motion, which can be described by explicit trajectories or formulas, complex deformation dynamics remain challenging to synthesize. We observe that a lack of physical reasoning for localizing dynamic areas allows irrelevant regions to dilute the model's attention, leading to generation failure. In this paper, we propose \textbf{DeforM}, a reasoning-guided image-to-video generation framework that directs the model's focus toward physics-critical regions. To reason and localize these critical regions, we introduce a VLM-guided physical reasoning module, DeforM-Reason, to identify target objects and generate spatial-temporal masks. For physical guidance, we develop two alternative strategies: DeforM-Free for training-free mechanism analysis, and DeforM-Injection as a powerful training-based generator. Experimental results demonstrate that DeforM improves the realism of generated deformation scenarios, outperforming baseline models in both visual quality and physical consistency. More information is available on our project page: \url{https://nemosunny.github.io/DeforM-ProjectPage/}.
\end{abstract}

\section{Introduction}
\label{sec:intro}

Video generation models have achieved significant advances, attracting growing attention for film production~\cite{Hunyuan, Wan, CogVideoX, LAVIE} and world modeling~\cite{WorldDreamer, he2025pretrainedvideogenerativemodels}. While such models provide realistic visuals for simulating virtual environments, these models do not guarantee underlying consistency with the laws of physics~\cite{PhysGen,VLIPP, PhysMotion}. 
Beyond improving visual quality, ensuring physical consistency is essential yet remains an unresolved challenge for practical video generators.

Despite various efforts~\cite{VLIPP,PISA,PhysRVG} to address this issue, existing approaches still face significant bottlenecks.
Specifically, simulator-based methods~\cite{PhysGen,PhysMotion,WonderPlay} require explicit scene reconstruction, which limits their scalability to diverse real-world environments. Alternatively, data-driven and reward-based methods~\cite{PhysHPO,WISA} are often constrained by limited physical data diversity and complex reward engineering. More recently, some approaches utilize visual large language models (VLMs)~\cite{qwen,qwen2.5,qwen3} to steer video generation towards physical plausibility~\cite{VLIPP,PhyRPR}, but are restricted to rigid-body dynamics.
Unlike rigid motion, deformation dynamics, such as melting, squeezing, and fracturing, involve high-dimensional transformations that are difficult to characterize through simple formulas or trajectory sequences. Furthermore, existing generation models struggle to reason about the physical properties of objects based on the visual and textual inputs, thus failing to focus on physics-critical regions and leading to physically implausible artifacts.

To this end, we propose \textbf{DeforM}, a reasoning-guided image-to-video generation framework that directs the video generation model's focus toward physics-critical regions to generate realistic deformation dynamics. To achieve this, two crucial questions must be addressed: \textit{how to leverage physical reasoning to precisely localize the regions where deformation dynamics occur} and \textit{how to effectively steer the model's attention toward these critical areas}. \Cref{fig:fig1} illustrates the motivation behind DeforM and shows representative results of our method.

\textbf{VLM-guided Physical Reasoning.}
To empower the video generation process with physical property reasoning and precise phenomena localization, we propose a VLM-guided physical reasoning module, called \textbf{DeforM-Reason}. This module acts as a cognitive agent by identifying the target objects most relevant to the deformation described in the text prompt and the initial image. We then employ an open-vocabulary object detector to ground these objects and generate a precise spatial-temporal mask. The mask is then aligned with video latents, providing concrete geometric priors to inform the model where to apply physical laws throughout the generation process.

\textbf{Spatial-temporal Mask Injection.}
To integrate physical priors into the generation process, we provide two alternative strategies for mask-guided modulation: \textbf{DeforM-Free} and \textbf{DeforM-Injection}. We initiate our study with DeforM-Free, a training-free approach that modulates cross-attention queries~\cite{attention} to concentrate generative capacity on the identified regions. This plug-and-play exploration reveals a fundamental observation: achieving physical consistency primarily requires the model to refocus its globally dispersed attention onto physics-critical areas. Building upon this insight, we further develop DeforM-Injection as a more powerful, training-based alternative to push the boundaries of deformation realism. It employs a dense mask modulator to inject high-dimensional features into both cross-attention and self-attention blocks, optimized via a specialized region-sensitive loss. While both methods independently enable physics-aware video generation, DeforM-Free unveils the underlying mechanism through step-by-step analysis, whereas DeforM-Injection provides a powerful generator for diverse and realistic deformation.

We summarize our contributions as follows:
1) We propose DeforM, a reasoning-guided image-to-video generation framework to generate realistic deformation dynamics by guiding the model's focus toward essential physical regions.
2) We introduce a VLM-guided physical reasoning module, DeforM-Reason, for precise spatial-temporal localization of deformation. Furthermore, we develop two alternative methods: DeforM-Free, which facilitates the analysis and validation of our core observations, and DeforM-Injection, a powerful training-based generator designed for diverse and realistic deformation dynamics.
3) Extensive experiments show that DeforM surpasses state-of-the-art (SOTA) methods in both visual quality and physical consistency, achieving physics-aware video generation.

\section{Related Work}
\label{sec:related_work}

\subsection{Physics-Aware Video Generation}
Achieving physics-aware video generation is an essential objective for practical applications, spurring numerous studies~\cite{PhysGen, WISA, VLIPP}. These approaches can be categorized by the source of physical information they incorporate. Some works, like PhysGen~\cite{PhysGen}, PhysMotion~\cite{PhysMotion}, and WonderPlay~\cite{WonderPlay}, rely on simulator-based frameworks to generate videos that adhere to physical laws. However, the requirement for building complex assets and the inherent complexity of their extended pipelines limit their efficiency. Another branch of research focuses on learning physical laws from large-scale data. For instance, WISA~\cite{WISA}, PISA Experiments~\cite{PISA}, and Force Prompting~\cite{Force_Prompt} attempt to acquire physical priors from a vast amount of real-world or simulated data. Furthermore, methods represented by PhysRVG~\cite{PhysRVG} and PhysHPO~\cite{PhysHPO} employ reinforcement learning to capture physical information. However, these methods rely heavily on the design of sophisticated reward functions and are prone to reward hacking. Lastly, a burgeoning category of methods leverages VLMs to guide video generation models in producing physically plausible videos. Specifically, VLIPP~\cite{VLIPP} and PhyRPR~\cite{PhyRPR} utilize VLMs to plan object trajectories for training-free video generation. However, relying on trajectory bounding boxes or simple formulas fails to capture fine-grained deformation, inherently limiting these approaches to rigid-body motions, such as collisions and free falls. Furthermore, current generation models primarily focus on visual pattern synthesis, lacking the explicit physical reasoning capabilities required to understand the physical dynamics of complex scenes.

\subsection{VLM-Guided Visual Generation}
VLMs exhibit robust multimodal understanding and interaction capabilities, which are leveraged to enhance the quality of visual generation and editing. Some studies synergize the generative power of diffusion models with the advanced reasoning and planning capabilities of VLMs. For image editing, R-Genie~\cite{R-Genie} utilizes VLM-based reasoning to address the lack of deep comprehension regarding user intentions and contextual reasoning. DiffSensei~\cite{DiffSensei} employs VLMs to decompose complex prompts into structured goals, ensuring precise alignment with multi-objective instructions. For video generation, VLIPP~\cite{VLIPP} and PhyRPR~\cite{PhyRPR} utilize VLMs to plan plausible trajectories for object motion. Alternatively, several approaches employ VLMs as evaluators to improve generation quality through closed-loop feedback. GenPilot~\cite{GenPilot} utilizes VLM-based visual question answering and image captioning for systematic error analysis, guiding iterative prompt optimization. Similarly, VIVA~\cite{VIVA} introduces the Edit-GRPO algorithm for video editing, which leverages VLMs to provide relative rewards based on instruction fidelity, content preservation, and aesthetic quality.

\section{DeforM Framework}

DeforM aims to generate realistic deformation videos that adhere to physical dynamics, covering melting, squeezing, fracturing, stretching, and slicing. We focus on the image-to-video (I2V) task of synthesizing the video $\mathbf{x}_{1:T}$ from an initial image $c_{img}$ and a textual prompt $c_{text}$, represented as:
\begin{equation}
(c_{img}, c_{text}) \xrightarrow{\text{DeforM}} \mathbf{x}_{1:T},
\end{equation}
where $T$ denotes the number of generated frames. Our primary objective is to ensure that the output video $\mathbf{x}_{1:T}$ is physically plausible, accurately capturing intricate physical evolutions described in $c_{text}$.

\begin{figure}[t]
    \centering
    \includegraphics[width=1.0\linewidth]{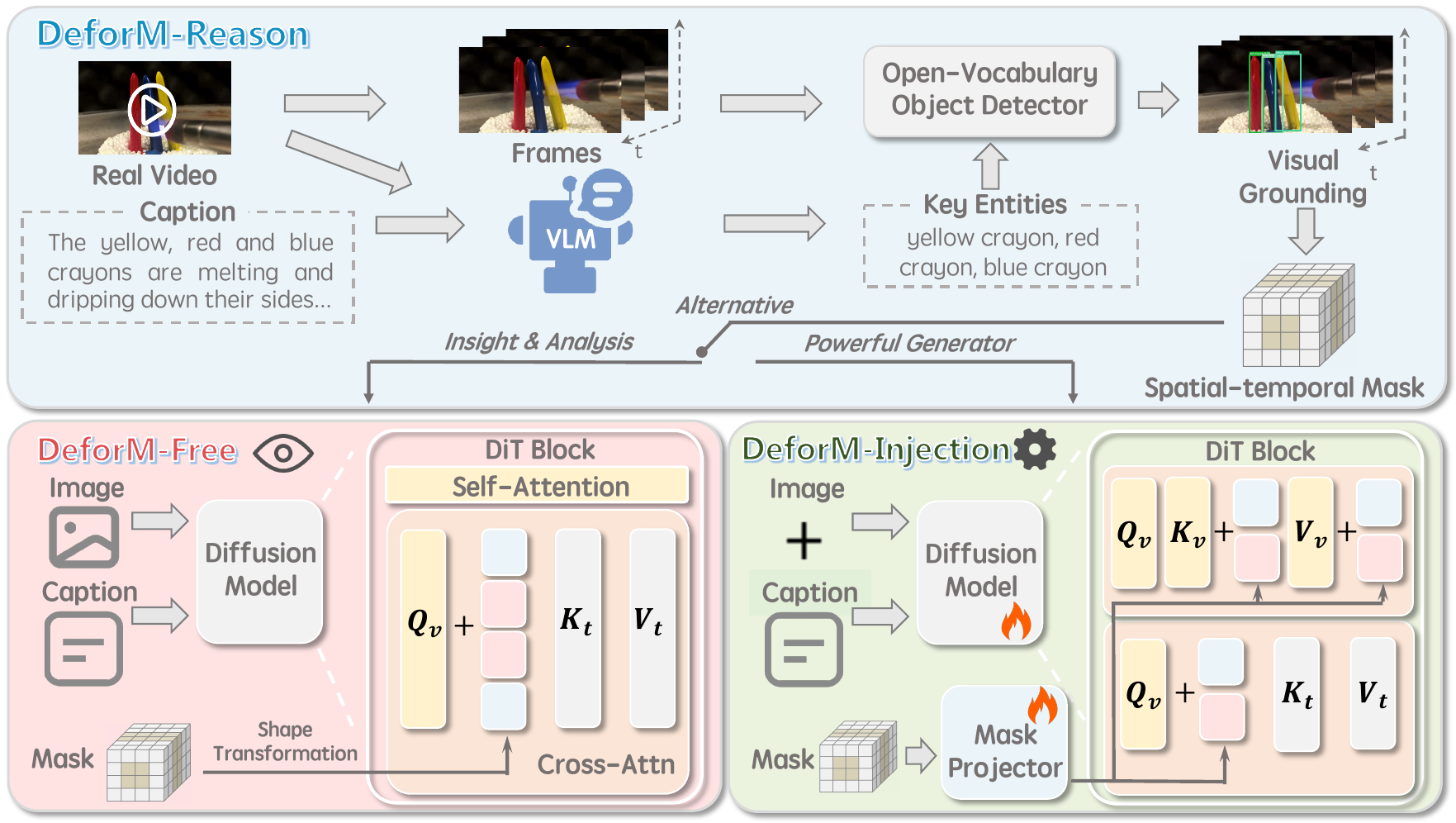}
    \caption{\textbf{Illustration of DeforM.} DeforM takes a text prompt and an initial image as inputs. The DeforM-Reason module first generates spatial-temporal masks by localizing physics-critical regions via VLM-guided reasoning. For physical guidance, we provide two alternative strategies: (1) DeforM-Free, a training-free method for mechanism analysis and observation validation. (2) DeforM-Injection, a powerful training-based generator for diverse and realistic deformation dynamics.}
    \label{fig:overview}
\end{figure}

As illustrated in~\cref{fig:overview}, DeforM comprises a reasoning module, DeforM-Reason, and two alternative strategies for the model's attention redirection: DeforM-Free and DeforM-Injection. Specifically, DeforM-Reason integrates VLM-guided physical reasoning to analyze the input text prompt $c_{text}$ and the initial image $c_{img}$ for precise localization of physics-critical regions (\cref{sec:DeforM-Reason}). To redirect the model's attention, DeforM-Free employs a training-free mechanism to incorporate physical priors into the inference process (\cref{sec:DeforM-Free}). This preliminary exploration reveals the significance of explicit reasoning and attention anchoring. Building upon these insights, the training-based DeforM-Injection is proposed to enhance deformation realism and scene diversity, synthesizing physics-aware videos (\cref{sec:DeforM-Injection}).

\subsection{Preliminary: Flow Matching for Video Generation}
\label{sec:preliminary}

Flow matching has emerged as a powerful paradigm for high-fidelity video generation due to its superior efficiency and sampling quality~\cite{pyramidal-flow-matching, efficient-flow-matching, flow-matching, Wan}. In this work, we build our DeforM framework upon a pre-trained flow-matching-based video generator. In the training stage, video flow matching models first map an input video $\mathbf{x}$ into a latent tensor $\mathbf{z}_0 = \mathcal{E}(\mathbf{x})$ via a pretrained encoder $\mathcal{E}(\cdot)$~\cite{LDM}. Instead of predicting noise, the model $\mathbf{v}_\theta$ learns to predict the velocity field governing the probability path between the data distribution and the noise distribution. We adopt the Optimal Transport (OT) path~\cite{flow-matching, albergo2023building}, where the intermediate state $\mathbf{z}_t$ at timestep $t\in[0,1]$ is constructed as:
\begin{equation}
    \mathbf{z}_t = (1-t)\mathbf{z}_0 + t\mathbf{z}_1,
\end{equation}
where $\mathbf{z}_1\sim \mathcal{N}(\mathbf{0},\mathbf{I})$ represents Gaussian noise. The model is trained to regress the ground-truth velocity $\mathbf{u}_t(\mathbf{z}_t \mid \mathbf{z}_0,\mathbf{z}_1)=\mathbf{z}_1-\mathbf{z}_0$ by minimizing:
\begin{equation}
    \mathcal{L}(\theta) = \mathbb{E}_{\mathbf{z}_0, \mathbf{z}_1, t, c} \bigl\| \mathbf{v}_\theta(\mathbf{z}_t,t,c)-(\mathbf{z}_1-\mathbf{z}_0) \bigr\|^2,
\end{equation}
where $c$ denotes conditions~\cite{cfg}, such as text and image prompts, and $t$ is sampled uniformly from $[0,1]$. In the inference stage, video generation is formulated as solving an Ordinary Differential Equation (ODE)~\cite{ODE}. Starting from a noise sample $\mathbf{z}_1 \sim \mathcal{N}(\mathbf{0},\mathbf{I})$, the model generates the latent video $\mathbf{z}_0$ by numerically integrating the predicted velocity field $\mathbf{v}_\theta$ from $t=1$ to $t=0$.

\subsection{DeforM-Reason: VLM-Guided Physical Reasoning}
\label{sec:DeforM-Reason}

A prerequisite for physics-aware video generation is the precise identification of where the physical laws should be applied. However, existing video generation models often lack such physical reasoning capabilities. Specifically, while a text prompt $c_{text}$ may specify deformation dynamics, the model often struggles to autonomously associate this semantic instruction with the corresponding pixels in the input image $c_{img}$. We observe that this semantic-spatial gap is a primary bottleneck. To bridge this gap, we propose DeforM-Reason, a VLM-guided physical reasoning module that acts as a cognitive agent, translating abstract physical instructions into concrete spatial-temporal constraints.

\begin{algorithm}[t]
\caption{DeforM-Reason: VLM-Guided Physical Reasoning}
\label{alg:deform_reason}
\begin{algorithmic}[1]
\REQUIRE Image $c_{img}$, prompt $c_{text}$, video length $T$, mode $\in \{\text{train, infer}\}$.
\ENSURE Latent-aligned mask $\mathbf{M}$.
\STATE $\mathcal{O}_{target} \leftarrow \textsc{Reason}(c_{img}, c_{text})$ \COMMENT{Identify target via VLM}
\IF{mode = train}
    \STATE $\{\mathbf{B}_i\}_{i=1}^T \leftarrow \textsc{Detect}(\mathcal{O}_{target}, \text{GT frames})$
\ELSE
    \STATE $\mathbf{B}_1 \leftarrow \textsc{Detect}(\mathcal{O}_{target}, c_{img})$
    \STATE $\{\mathbf{B}_i\}_{i=2}^T \leftarrow \textsc{Plan}(\mathbf{B}_1, c_{text})$ \COMMENT{VLM-based temporal extrapolation}
\ENDIF
\STATE $\mathbf{M}_{mask} \leftarrow \mathbf{0} \in \{0,1\}^{T \times H \times W}$
\FOR{$i=1$ \TO $T$}
    \STATE $\mathbf{M}_{mask}[i, (x, y) \in \mathbf{B}_i] \leftarrow 1$
\ENDFOR
\STATE $\mathbf{M} \leftarrow \textsc{Downsample}(\mathbf{M}_{mask})$ \COMMENT{Align with latent shape $(T', H', W')$}
\RETURN $\mathbf{M}$
\end{algorithmic}
\end{algorithm}

As illustrated in \textbf{Algorithm \ref{alg:deform_reason}}, given $c_{img}$ and $c_{text}$, we first employ a VLM for physical reasoning to identify the object most relevant to the described deformation, denoted as $\mathcal{R}(c_{img}, c_{text}) \rightarrow \mathcal{O}_{target}$, where $\mathcal{O}_{target}$ is the semantic identifier of the target object. Subsequently, an open-vocabulary object detector is utilized to establish spatial grounding.
The grounding process varies between the training and inference stages:
\begin{itemize}[label=\textbullet, leftmargin=*]
\item \textbf{Training Stage.} We utilize the ground-truth video sequences to obtain precise bounding box sequences. The detector identifies the target object $\mathcal{O}_{target}$ across all frames $i \in \{1, \dots, T\}$, yielding a sequence of bounding boxes $\{\mathbf{B}_i\}_{i=1}^T$.
\item \textbf{Inference Stage.} Since the future frames are unavailable during inference, we leverage the VLM to reason about the temporal localization of the object. The detector first locates the object in $c_{img}$ to obtain $\mathbf{B}_1$. The VLM then performs temporal planning to extrapolate the first box into a sequence of spatial-temporal constraints $\{\mathbf{B}_i\}_{i=2}^T$, accounting for the expected deformation described in $c_{text}$.
\end{itemize}
Finally, we construct a binary spatial-temporal mask $\mathbf{M} \in \{0, 1\}^{T \times H \times W}$ by setting the interior of each $\mathbf{B}_i$ to 1. To align with the generative process, $\mathbf{M}$ is downsampled to the dimensions of the video latents $\mathbf{z} \in \mathbb{R}^{T' \times H' \times W'}$, where $T', H', W'$ denote the temporal and spatial dimensions in the latent space. The resulting mask serves as an explicit spatial-temporal prior, effectively guiding the model to concentrate its generative capacity on regions where deformation dynamics occur.

By transforming the implicit physical potential of an image into an explicit spatial-temporal mask, we provide the foundation for the subsequent training-free attention modulation and training-based methods, effectively guiding the model to focus on the regions where physics matters.

\subsection{DeforM-Free: Spatial-Temporal Attention Modulation}
\label{sec:DeforM-Free}
\begin{figure}[t]
    \centering
    \includegraphics[width=1.0\linewidth]{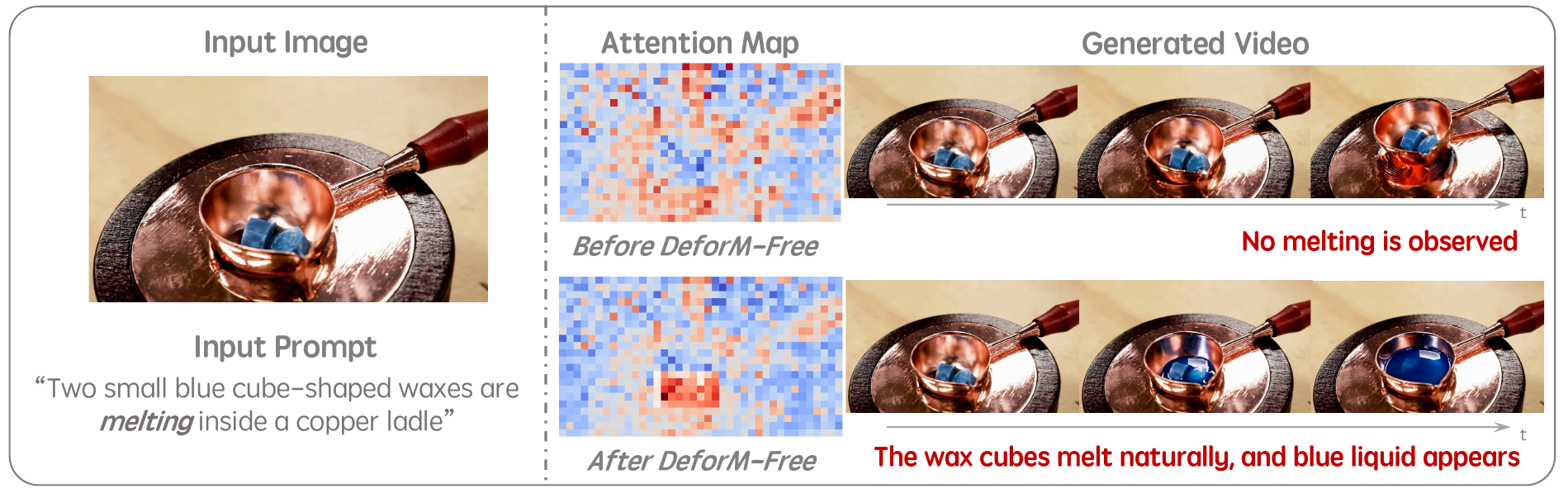}
    \caption{\textbf{Motivation and Analysis of DeforM-Free.}(Top) Base models exhibit a semantic-spatial gap, where cross-attention for the keyword \textit{melting} is spatially diffused, leading to physical inconsistency. (Bottom) By anchoring attention via DeforM-Free, the attention map becomes more active on the target region, resulting in physically plausible deformation dynamics.}
    \label{fig:fig_3}
\end{figure}

The primary challenge in physics-aware video generation lies in the discrepancy between linguistic physical priors and localized visual generation. In this section, our empirical analysis reveals that the failure of base models to generate realistic deformation stems from this semantic-spatial gap. As illustrated in \cref{fig:fig_3}, the base model fails to concentrate the \textit{melting} instruction on the target object regions, resulting in physically inconsistent generation where the melting occurs in irrelevant areas.

By visualizing the cross-attention maps~\cite{attention} for the keyword \textit{melting}, we observe that the attention signals are diluted across the entire frame. This observation confirms that the model struggles to autonomously associate abstract physical keywords with the corresponding spatial pixels. This insight serves as the foundation for our proposed DeforM-Free, as a training-free modulation method designed to bridge this gap by explicitly anchoring the model's attention.

We leverage the spatial-temporal mask $\mathbf{M} \in \{0,1\}^{T' \times H' \times W'}$ obtained from DeforM-Reason as a prior to modulate the attention mechanism. Let $\mathbf{z}$ be the video latent features and $\mathbf{e}_{text}$ be the text prompt embedding. In the cross-attention blocks, the queries $\mathbf{Q}$, keys $\mathbf{K}$, and values $\mathbf{V}$ are calculated as:
\begin{equation}
\mathbf{Q} = \mathbf{z}\mathbf{W}_q, \quad \mathbf{K} = \mathbf{e}_{text}\mathbf{W}_k, \quad \mathbf{V} = \mathbf{e}_{text}\mathbf{W}_v,
\end{equation}
where $\mathbf{W}_q$, $\mathbf{W}_k$, $\mathbf{W}_v$ are the projection matrices. To reinforce the coupling between the physical keywords and target objects, we apply a modulation factor $\alpha$ to the queries within the masked regions. We first denote $\hat{\mathbf{M}} \in \{0, 1\}^{S}$ as the flattened and interpolated mask aligned with the latent sequence length $S$, where $S = T' \times H' \times W'$. The modulated queries $\tilde{\mathbf{Q}}$ are defined as:
\begin{equation}
\tilde{\mathbf{Q}}_j = \mathbf{Q}_j \cdot [1 + \hat{\mathbf{M}}_j \cdot (\alpha - 1)],
\end{equation}
where $j$ is the token index and $\alpha > 1$ is the focus scale. By scaling $\mathbf{Q}$ in the masked area, we effectively amplify the attention scores $\text{Softmax}(\frac{\tilde{\mathbf{Q}}\mathbf{K}^T}{\sqrt{d}})\mathbf{V}$, forcing the model to prioritize the target regions.

As shown in~\cref{fig:fig_3}, after applying DeforM-Free, the attention map for \textit{melting} becomes more concentrated on the masked region. The generated video exhibits accurate deformation dynamics, where the wax melts correctly within the boundaries. This validation demonstrates that guiding the model toward physics-critical regions is essential for capturing deformation dynamics. By providing the spatial bias without altering pre-trained weights, DeforM-Free establishes the empirical basis for our subsequent training-based method, DeforM-Injection.

\subsection{DeforM-Injection: Training-based Method}
\label{sec:DeforM-Injection}

While the DeforM-Free method effectively guides the model's attention, it is inherently limited by the base model's pre-existing knowledge of physical laws. If the base model lacks a sufficient internal representation of specific deformation dynamics, simply scaling attention scores may result in artifacts or physically inconsistent dynamics. To overcome these limitations, we propose DeforM-Injection, a training-based framework that explicitly injects spatial-temporal priors into the latent space for complex deformation dynamics generation. 

To represent the precise physics-critical regions, we design a Dense Mask Modulator $\mathcal{F}_{\phi}$, serving as an encoder for the spatial-temporal masks. Specifically, given the binary mask $\mathbf{M} \in \{0, 1\}^{T' \times H' \times W'}$ obtained from DeforM-Reason, we first apply a series of 3D convolutional layers to extract hierarchical features while downsampling the mask to match the dimensions of the video latents. The modulator projects the mask into a high-dimensional feature space, generating three decoupled modulation streams: $\Delta \mathbf{k}_{\text{self}}$, $\Delta \mathbf{v}_{\text{self}}$, and $\Delta \mathbf{q}_{\text{cross}}$. To ensure training stability and preserve the foundation model's generation capability, we employ a zero-initialization strategy for the final linear layers of the modulator, allowing the network to gradually learn the required offsets.

The core idea of DeforM-Injection lies in the synchronized modulation of both self-attention and cross-attention blocks. Unlike DeforM-Free, which is restricted to enhancing semantic anchoring via training-free cross-attention scaling, this synchronized strategy enables the model to jointly refine external semantic retrieval and internal spatial-temporal consistency through learned feature offsets. We first extend the semantic anchoring concept from DeforM-Free to the training stage. We inject the learned modulation into the cross-attention queries $\mathbf{Q}_{\text{cross}}$, which represent the latent tokens' active request for physical information from the text prompt. By augmenting $\mathbf{Q}_{\text{cross}}$, the masked regions are empowered to more effectively retrieve physical knowledge from the text embedding. While cross-attention blocks provide semantic guidance, self-attention blocks govern the internal spatial-temporal consistency of the video. We inject mask features into the self-attention keys $\mathbf{K}_{\text{self}}$ and values $\mathbf{V}_{\text{self}}$ within the identified regions. This allows the model to learn localized physical correlations, facilitating the generation of complex and physically plausible deformation dynamics. The modulated attention can be defined as follows:
\begin{equation}
\begin{aligned}
    \mathbf{K}'_{\text{self}} &= \mathbf{K}_{\text{self}} + \Delta \mathbf{k}_{\text{self}}, \\
    \mathbf{V}'_{\text{self}} &= \mathbf{V}_{\text{self}} + \Delta \mathbf{v}_{\text{self}}, \\
    \mathbf{Q}'_{\text{cross}} &= \mathbf{Q}_{\text{cross}} + \Delta \mathbf{q}_{\text{cross}},
\end{aligned}
\end{equation}
where $\Delta \mathbf{k}_{\text{self}}, \Delta \mathbf{v}_{\text{self}}, \Delta \mathbf{q}_{\text{cross}} = \mathcal{F}_{\phi}(\mathbf{M})$.

We optimize both the Dense Mask Modulator and a set of Low-Rank Adaptation (LoRA)~\cite{lora} weights integrated into the transformer backbone. To further prioritize the fidelity of physical deformations, we define a composite loss function. In addition to the standard flow matching loss $\mathcal{L}_{\text{flow}}$ applied to the entire latent sequence, we introduce a region-sensitive loss $\mathcal{L}_{\text{mask}}$ that assigns higher weights to the reconstruction error within physics-critical regions defined by $\mathbf{M}$. The total training objective is:
\begin{equation}
\mathcal{L}_{total} = (1 - \lambda_{mask}) \mathcal{L}_{flow} + \lambda_{mask} \mathcal{L}_{mask},
\end{equation}
where $\lambda_{mask}$ is a balancing hyperparameter. By focusing the optimization on the physics-critical regions, DeforM-Injection enables the model to bridge the gap between abstract physical concepts and concrete pixel-level dynamics, producing videos that are both visually natural and physically plausible.

\section{Experiments}
This section presents the experimental validation of our proposed methods across multiple dimensions. Specifically, \cref{sec:implementations} provides the implementation details for both DeforM-Free and DeforM-Injection. \cref{sec:evaluation_settings} introduces the datasets, baselines and the design of evaluation metrics. To demonstrate the effectiveness of our methods, \cref{sec:comparisons} reports both qualitative and quantitative results compared with SOTA models. The rationality and performance of our designed modules are further analyzed in \cref{sec:ablation_study}.

\subsection{Implementations}
\label{sec:implementations}
We adopt Wan2.2-TI2V-5B~\cite{Wan} as the base image-to-video generation model. To achieve a balance between inference speed and performance, we employ Qwen2.5-VL-7B~\cite{qwen2.5} as the physical reasoning agent and LLMDet~\cite{llmdet} as the open-vocabulary object detector for DeforM-Reason. DeforM-Free is implemented as a training-free strategy by modulating the cross-attention blocks during the inference stage. We set the focus scale $\alpha$ to 1.4 to provide precise spatial-temporal guidance while maintaining generation quality.

For DeforM-Injection, we fine-tune Wan2.2-TI2V-5B using LoRA with a rank of 256. The training process utilizes a batch size of 64 with a learning rate of $1 \times 10^{-6}$ for the LoRA weights and a higher rate of $2 \times 10^{-6}$ for the modulator. The balancing hyperparameter $\lambda_{mask}$ is set to 0.3 to prioritize the reconstruction of physics-critical regions.

\subsection{Evaluation Settings}
\label{sec:evaluation_settings}
\begin{itemize}[label={}, leftmargin=0pt, nosep]
\item \textbf{Datasets.} We conduct our experiments using a specialized subset of the WISA-80K~\cite{WISA} dataset. To ensure training quality, we select approximately 6k high-quality deformation dynamics samples, including melting, squeezing, fracturing, and other scenarios. Each sample consists of a ground-truth video captured in real-world environments with a corresponding text caption describing the physical action. For evaluation, we utilize a separate test set comprising 320 samples across diverse scenarios for visual quality and randomly select 160 samples for physical consistency evaluation. These evaluation samples are excluded from the training set across diverse deformation categories.

\item \textbf{Baselines.} To evaluate the performance of our framework, we compare it against several open-source SOTA models that have demonstrated robust image-to-video generation capabilities. These competitive baselines include Hunyuan-I2V~\cite{Hunyuan}, CogVideoX1.5-I2V~\cite{CogVideoX}, Wan2.2-14B-I2V~\cite{Wan} and MAGI-1~\cite{magi}. We selected these models for their outstanding video generation performance and widespread usage.

\item \textbf{Evaluation Metrics.}
For a comprehensive evaluation, we assess the generated results across both visual quality and physical consistency dimensions. Regarding visual quality, we adopt VBench~\cite{vbench} as the foundational benchmark following previous works~\cite{PhyRPR, PhysRVG}. Considering the unique nature of deformation dynamics where foreground objects undergo significant structural changes and may affect the background, we exclude \textit{subject consistency} and \textit{background consistency} as these attributes are inherently compromised during such processes. Instead, we retain six other metrics that comprehensively reflect video quality: \textit{motion smoothness}, \textit{temporal flickering}, \textit{dynamic degree}, \textit{aesthetic quality}, \textit{imaging quality} and the aggregate overall \textit{quality score}.

\hspace{2em}For physical consistency, we utilize Qwen3-VL-32B~\cite{qwen3} as a powerful evaluator, considering its exceptional multimodal physical evaluation capabilities. The generated videos and corresponding captions are provided as inputs to score the results across three distinct dimensions using a scale from 1 to 5. The first dimension is \textit{Physical Commonsense} (PC), which evaluates the adherence to the general physics of the entire video. The second dimension is \textit{Semantic Adherence} (SA), assessing how well the video follows the textual instructions. The third dimension is \textit{Local Deformation Fidelity} (LDF), which focuses on the accuracy and realism of local deformation dynamics. For each dimension, the score ranges from 1 representing \textit{Poor} and 5 representing \textit{Excellent}, providing a detailed quantitative measure of physical realism. We define the \textit{Physical Score} as the arithmetic mean of these three dimensions, providing a comprehensive assessment of the overall physical plausibility.
\end{itemize}

\subsection{Comparisons}
\label{sec:comparisons}

\begin{table*}[t]
\centering
\caption{Quantitative comparison of DeforM-Injection with state-of-the-art image-to-video generation models. Blue and yellow shaded columns represent the aggregate scores for visual quality and physical consistency, respectively. The best results are highlighted in bold.}
\label{tab:main_results}
\resizebox{\textwidth}{!}{
\begin{tabular}{l ccccc >{\columncolor{hlblue}}c ccc >{\columncolor{hlyellow}}c}
\toprule
\rowcolor{white}
& \multicolumn{6}{c}{\textbf{Visual Quality}} & \multicolumn{4}{c}{\textbf{Physical Consistency}} \\
\cmidrule(lr){2-7} \cmidrule(lr){8-11}
\rowcolor{white}
Method & \begin{tabular}[c]{@{}c@{}}Temporal \\ Flickering $\uparrow$\end{tabular} & \begin{tabular}[c]{@{}c@{}}Motion \\ Smoothness $\uparrow$\end{tabular} & \begin{tabular}[c]{@{}c@{}}Dynamic \\ Degree $\uparrow$\end{tabular} & \begin{tabular}[c]{@{}c@{}}Aesthetic \\ Quality $\uparrow$\end{tabular} & \begin{tabular}[c]{@{}c@{}}Imaging \\ Quality $\uparrow$\end{tabular} & \begin{tabular}[c]{@{}c@{}}Quality \\ Score $\uparrow$\end{tabular} & PC $\uparrow$ & SA $\uparrow$ & LDF $\uparrow$ & \begin{tabular}[c]{@{}c@{}}Physical \\ Score $\uparrow$\end{tabular} \\
\midrule
CogVideoX1.5-I2V         & 98.25  & 98.84  & 48.32  & 45.68  & 65.92  & 71.40  & 4.07  & 3.62  & 3.24  & 3.64  \\
Hunyuan-I2V              & \textbf{99.30}  & \textbf{99.54}  & 27.74  & 46.15  & 64.80  & 67.50  & 3.86  & 3.34  & 3.01  & 3.40  \\
Wan2.2-I2V-14B           & 97.42  & 98.34  & 46.07  & \textbf{46.61}  & 67.47  & 71.18  & 4.28  & 3.91  & 3.65  & 3.95  \\
MAGI-1              & 99.09  & 99.41  & 28.03  & 45.97  & 63.79  & 67.26  & 3.93  & 3.39  & 3.01  & 3.44  \\
\midrule
\textbf{DeforM-Injection (Ours)} & 97.24  & 98.40  & \textbf{54.74}  & \textbf{46.61}  & \textbf{68.84}  & \textbf{73.17}  & \textbf{4.59}  & \textbf{4.46}  & \textbf{4.16}  & \textbf{4.40}  \\
\bottomrule
\end{tabular}
}
\end{table*}

\begin{figure}[t]
    \centering
    \includegraphics[width=1.0\linewidth]{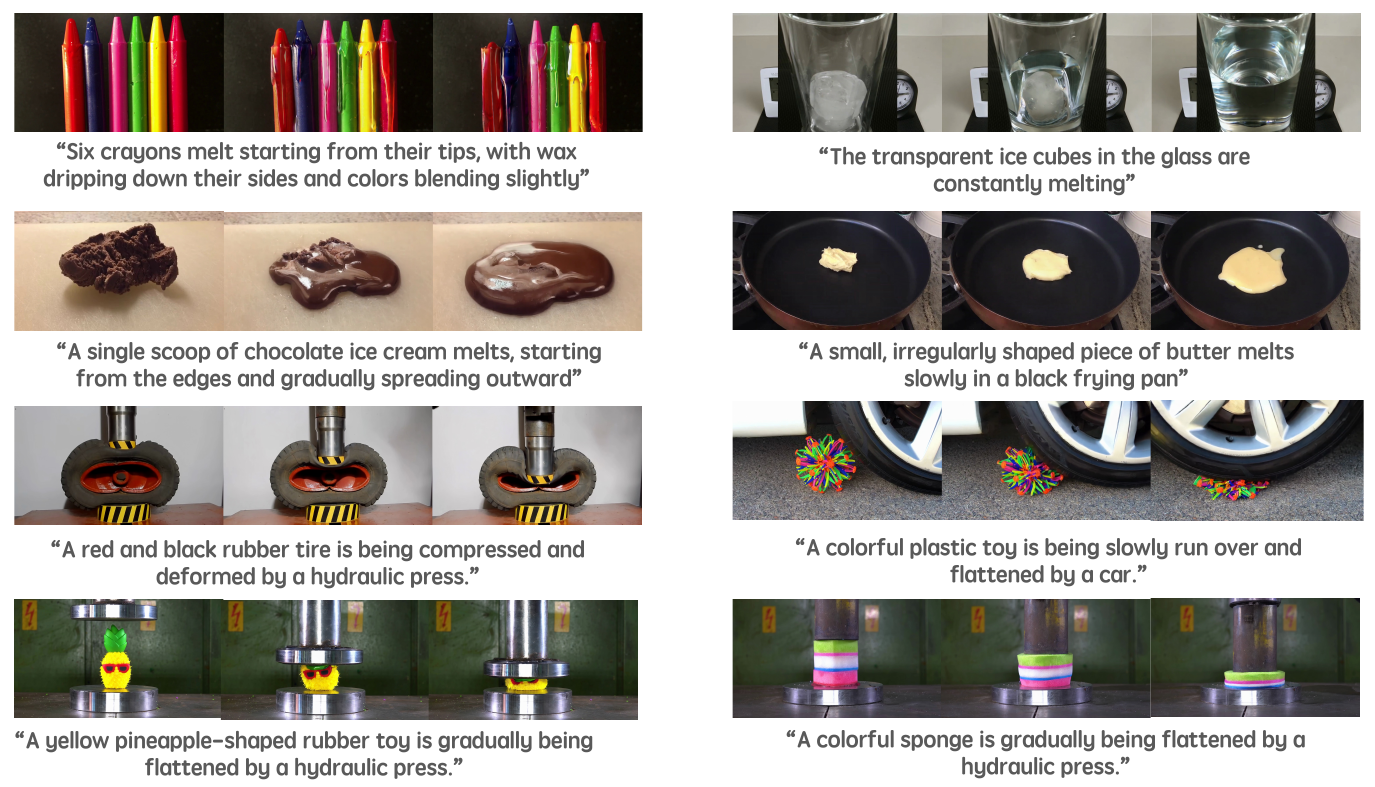}
    \caption{\textbf{Visualization results.} Visualization of diverse deformation scenarios generated by DeforM-Injection.}
    \label{fig:visual_results}
\end{figure}

\subsubsection{Quantitative Results.}
Summarized in~\cref{tab:main_results}, the quantitative comparison results demonstrate the superior performance of our proposed method in both visual quality and physical consistency. For visual quality, our method achieves the highest overall quality score and demonstrates a significant advantage in the dynamic degree. Although some baselines, like Hunyuan-I2V, show high performance in temporal flickering and motion smoothness, they tend to generate nearly static scenes with minimal movement. In contrast, our approach generates obvious and realistic deformation dynamics with competitive temporal consistency and motion smoothness. Regarding physical consistency, our method shows superior results across all dimensions with the most notable advantages in LDF. These results demonstrate the effectiveness of the integration of VLM-guided physical reasoning and spatial-temporal mask injection.

\subsubsection{Qualitative Results.}
\begin{figure}
    \centering
    \includegraphics[width=1.0\linewidth]{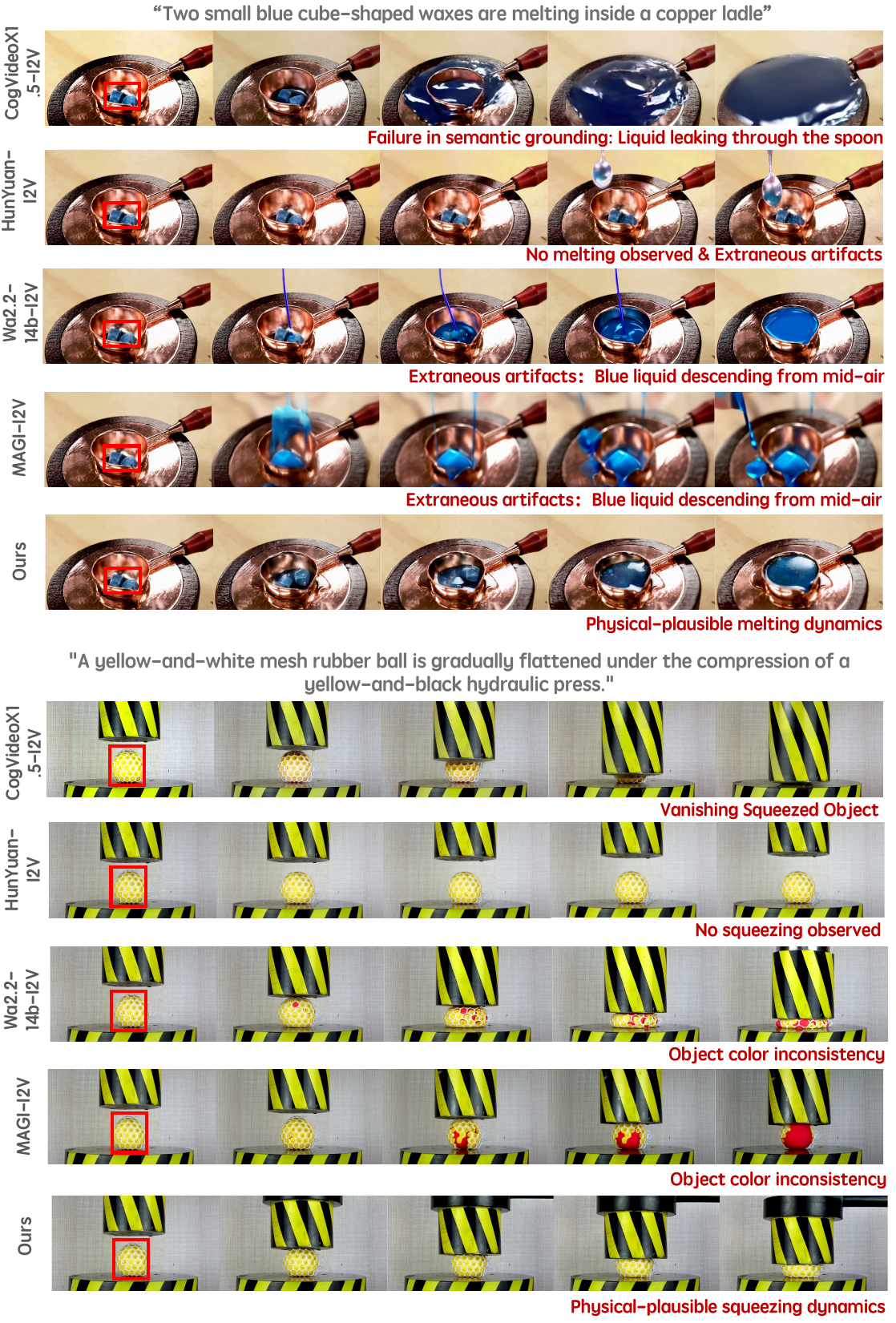}
    \caption{\textbf{Qualitative comparisons.} DeforM presents superior physical plausibility and vivid local deformation dynamics. The red bounding boxes indicate the physics-critical regions.}
    \label{fig:com_visual}
\end{figure}

In this section, we present qualitative results to demonstrate the generation quality and physical plausibility of our method. \cref{fig:visual_results} presents diverse deformation scenarios reflecting the physical consistency and dynamics achieved by our work. Furthermore, \cref{fig:com_visual} provides a visual comparison between DeforM and baselines across two representative deformation types involving melting and squeezing. In the melting examples, baseline models suffer from various limitations such as poor object understanding, failure to generate the melting effect, or extraneous artifacts. Regarding the squeezing scenarios, the baseline models exhibit significant issues including disappearing objects, a lack of squeezing dynamics, and inconsistent object colors. In contrast, our method demonstrates superior physical plausibility and vivid local deformation dynamics, indicating the effectiveness of our framework.

\subsection{Ablation Studies}
\label{sec:ablation_study}
To evaluate the effectiveness of each component, we conduct an ablation study against the vanilla Wan2.2-I2V-5B and standard supervised fine-tuning (SFT). As shown in~\cref{tab:ablation_results}, the vanilla model and SFT method struggle to generate accurate deformation dynamics with lower dynamic degree and physical scores. To assess the impact of DeforM-Reason, we replace our VLM-guided masks with randomly generated spatial-temporal masks. The performance degradation confirms that precise localization of physics-critical regions is essential for effective guidance. Notably, DeforM-Free achieves strong performance in physical plausibility without any additional training, even surpassing the SFT method.

Furthermore, we independently modulate cross-attention and self-attention, denoted as DeforM-Cross and DeforM-Self, respectively, as shown in~\cref{tab:attention_ablation}. We observe that cross-attention modulation, DeforM-Cross, benefits semantic adherence and physical commonsense by aligning deformation information with spatial-temporal regions. Injection into self-attention, DeforM-Self, contributes more to temporal smoothness and imaging quality. DeforM-Injection achieves the highest performance, especially in LDF and overall physical score.

\begin{table*}[!t]
\centering
\caption{Ablation study of different configurations and baselines. Blue and yellow shaded columns represent the aggregate scores for visual quality and physical consistency, respectively. The best results are highlighted in bold.}
\label{tab:ablation_results}
\resizebox{\textwidth}{!}{
\begin{tabular}{l ccccc >{\columncolor{hlblue}}c ccc >{\columncolor{hlyellow}}c}
\toprule
\rowcolor{white}
& \multicolumn{6}{c}{\textbf{Visual Quality}} & \multicolumn{4}{c}{\textbf{Physical Consistency}} \\
\cmidrule(lr){2-7} \cmidrule(lr){8-11}
\rowcolor{white}
Method & \begin{tabular}[c]{@{}c@{}}Temporal \\ Flickering $\uparrow$\end{tabular} & \begin{tabular}[c]{@{}c@{}}Motion \\ Smoothness $\uparrow$\end{tabular} & \begin{tabular}[c]{@{}c@{}}Dynamic \\ Degree $\uparrow$\end{tabular} & \begin{tabular}[c]{@{}c@{}}Aesthetic \\ Quality $\uparrow$\end{tabular} & \begin{tabular}[c]{@{}c@{}}Imaging \\ Quality $\uparrow$\end{tabular} & \begin{tabular}[c]{@{}c@{}}Quality \\ Score $\uparrow$\end{tabular} & PC $\uparrow$ & SA $\uparrow$ & LDF $\uparrow$ & \begin{tabular}[c]{@{}c@{}}Physical \\ Score $\uparrow$\end{tabular} \\
\midrule
Vanilla Wan2.2    & \textbf{98.12}  & \textbf{98.85}  & 42.61  & 46.28  & 67.40  & 70.65  & 3.97  & 3.31  & 3.19  & 3.49  \\
SFT                & 97.71  & 98.54  & 44.77  & 45.99  & 67.65  & 70.93  & 4.19  & 3.69  & 3.41  & 3.76  \\

W/O DeforM-Reason & 97.70  & 98.62  & 47.58     & 45.60     & 67.52     & 71.40     & 4.22    & 3.79    & 3.45    & 3.82    \\
\midrule
\textbf{DeforM-Free}       & 97.21  & 98.27  & 51.57  & 46.47  & 66.41  & 71.99  & 4.28  & 3.86  & 3.53  & 3.89  \\

\textbf{DeforM-Injection} & 97.24  & 98.40  & \textbf{54.74}  & \textbf{46.61}  & \textbf{68.84}  & \textbf{73.17}  & \textbf{4.59}  & \textbf{4.46}  & \textbf{4.16}  & \textbf{4.40}  \\
\bottomrule
\end{tabular}
}
\end{table*}

\begin{table*}[!t]
\centering
\caption{Ablation study of different attention injection mechanisms. Blue and yellow shaded columns represent the aggregate scores for visual quality and physical consistency, respectively. The best results are highlighted in bold.}
\label{tab:attention_ablation}
\resizebox{\textwidth}{!}{
\begin{tabular}{l ccccc >{\columncolor{hlblue}}c ccc >{\columncolor{hlyellow}}c}
\toprule
\rowcolor{white}
& \multicolumn{6}{c}{\textbf{Visual Quality}} & \multicolumn{4}{c}{\textbf{Physical Consistency}} \\
\cmidrule(lr){2-7} \cmidrule(lr){8-11}
\rowcolor{white}
Method & \begin{tabular}[c]{@{}c@{}}Temporal \\ Flickering $\uparrow$\end{tabular} & \begin{tabular}[c]{@{}c@{}}Motion \\ Smoothness $\uparrow$\end{tabular} & \begin{tabular}[c]{@{}c@{}}Dynamic \\ Degree $\uparrow$\end{tabular} & \begin{tabular}[c]{@{}c@{}}Aesthetic \\ Quality $\uparrow$\end{tabular} & \begin{tabular}[c]{@{}c@{}}Imaging \\ Quality $\uparrow$\end{tabular} & \begin{tabular}[c]{@{}c@{}}Quality \\ Score $\uparrow$\end{tabular} & PC $\uparrow$ & SA $\uparrow$ & LDF $\uparrow$ & \begin{tabular}[c]{@{}c@{}}Physical \\ Score $\uparrow$\end{tabular} \\
\midrule
DeforM-Cross     & 97.37  & 98.43  & 53.23  & 46.42  & 67.74  & 72.64  & 4.42  & 4.01  & 3.83  & 4.09  \\
DeforM-Self      & \textbf{97.68}  & \textbf{98.62}  & 53.16  & 46.50  & 68.21  & 72.83  & 4.22  & 3.73  & 3.33  & 3.76  \\
\midrule
\textbf{DeforM-Injection} & 97.24  & 98.40  & \textbf{54.74}  & \textbf{46.61}  & \textbf{68.84}  & \textbf{73.17}  & \textbf{4.59}  & \textbf{4.46}  & \textbf{4.16}  & \textbf{4.40}  \\
\bottomrule
\end{tabular}
}
\end{table*}

\section{Conclusion}
In this paper, we propose DeforM, a reasoning-guided image-to-video framework for physics-aware deformation synthesis. To enable physical reasoning in deformation generation, we introduce a VLM-guided module to analyze object physical properties in multimodal inputs and ground physics-critical regions as spatial-temporal masks. Furthermore, we develop training-free and training-based mask injection strategies to steer the model's generation focus. Extensive experiments demonstrate the effectiveness of our framework, especially in local deformation fidelity and dynamic degree. Our work highlights the importance of incorporating physical reasoning and spatial-temporal physical priors into generative frameworks, providing a promising pathway for physics-aware video generation.

\textit{Limitations and Future Work.} Despite promising results, DeforM remains limited by the localization accuracy of the DeforM-Reason module, which may produce suboptimal spatial-temporal masks in complex scenes. Moreover, as DeforM builds on LoRA fine-tuning of a foundation model, its generation quality and scene diversity are partly constrained by the base model. These limitations may be alleviated by future advances in grounding, reasoning, and foundation video generation.


\section*{Acknowledgements}
This work is supported by Shanghai Artificial Intelligence Laboratory. We gratefully acknowledge the support and research resources provided by Shanghai Artificial Intelligence Laboratory throughout this project.

%
%
\bibliographystyle{splncs04}
\bibliography{main}

\clearpage
\appendix
\renewcommand{\thesection}{\Alph{section}}
\renewcommand{\thesubsection}{\thesection.\arabic{subsection}}

\section{Details of Dataset Construction}
\subsection{Foundational Dataset: WISA-80K}
We adopt WISA-80K as the foundational dataset for constructing our high-quality dataset for deformation dynamics. The selection of WISA-80K is motivated by its comprehensive coverage of real-world physical phenomena across a vast range of categories. As an open-source large-scale dataset, it provides a diverse set of scenarios that are essential for training models to understand complex deformation and physical interactions. 

\begin{figure}
    \centering
    \includegraphics[width=1.0\linewidth]{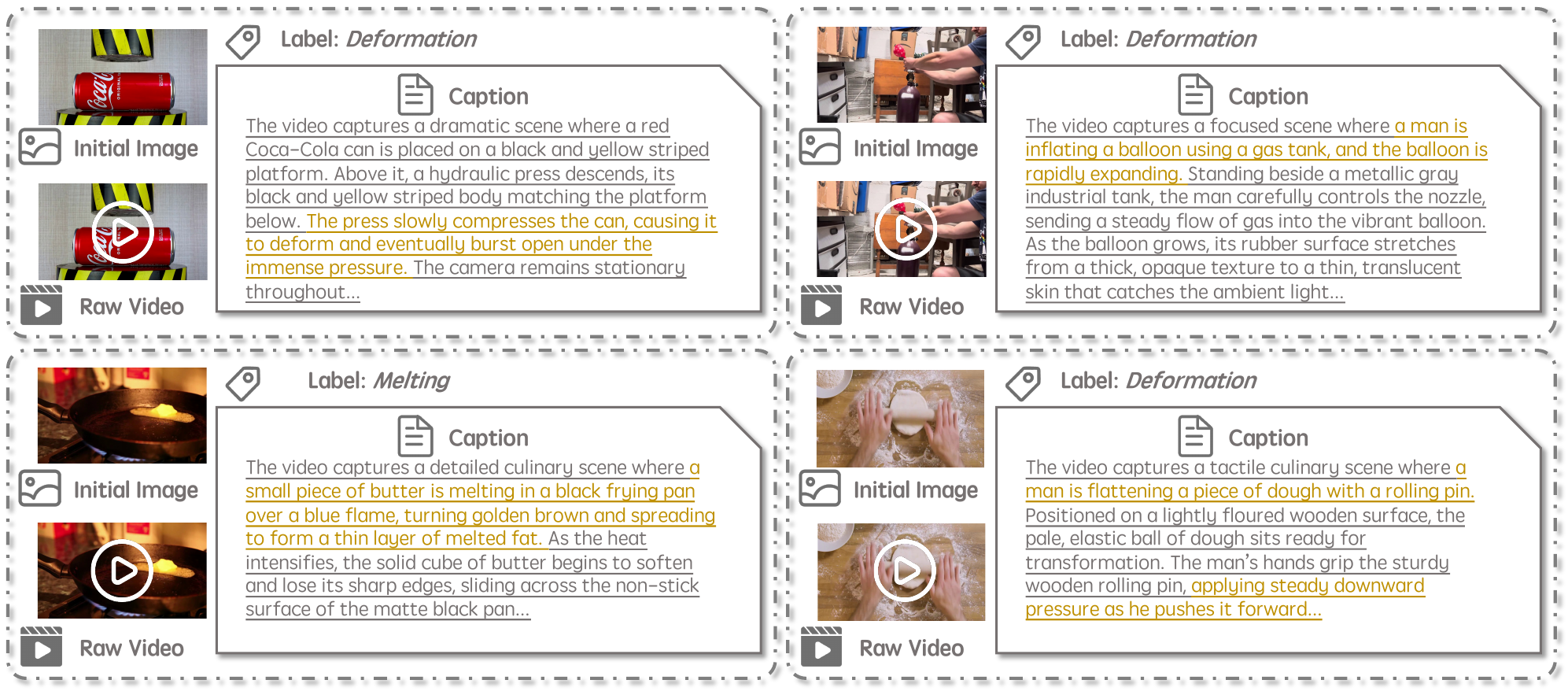}
    \caption{\textbf{Example of Extracted WISA-80K Data.} Visualization of raw WISA-80K data, including initial images, raw videos, labels and captions.}
    \label{fig:sm_fig1}
\end{figure}

In WISA-80K, each sample comprises a raw video with its corresponding metadata, such as captions, labels, and FPS. For our task, we selectively extract the video streams, the descriptive captions and the labels associated with each video. Specifically, the captions provide textual descriptions of the visual content, and the labels categorize the specific types of object dynamics in the corresponding video. Furthermore, the first frame of each video is preserved as the initial image, serving as the foundational visual condition for our subsequent data construction as shown in~\cref{fig:sm_fig1}.

\subsection{Dataset Filtering and Refinement}
To align the dataset with our focus on complex deformation dynamics, we selected the most suitable subsets from WISA-80K labeled as \textit{melting} and \textit{deformation}. These categories encompass a diverse range of deformation phenomena, including melting, squeezing, stretching, slicing, and fracturing.

\begin{figure}
    \centering
    \includegraphics[width=1.0\linewidth]{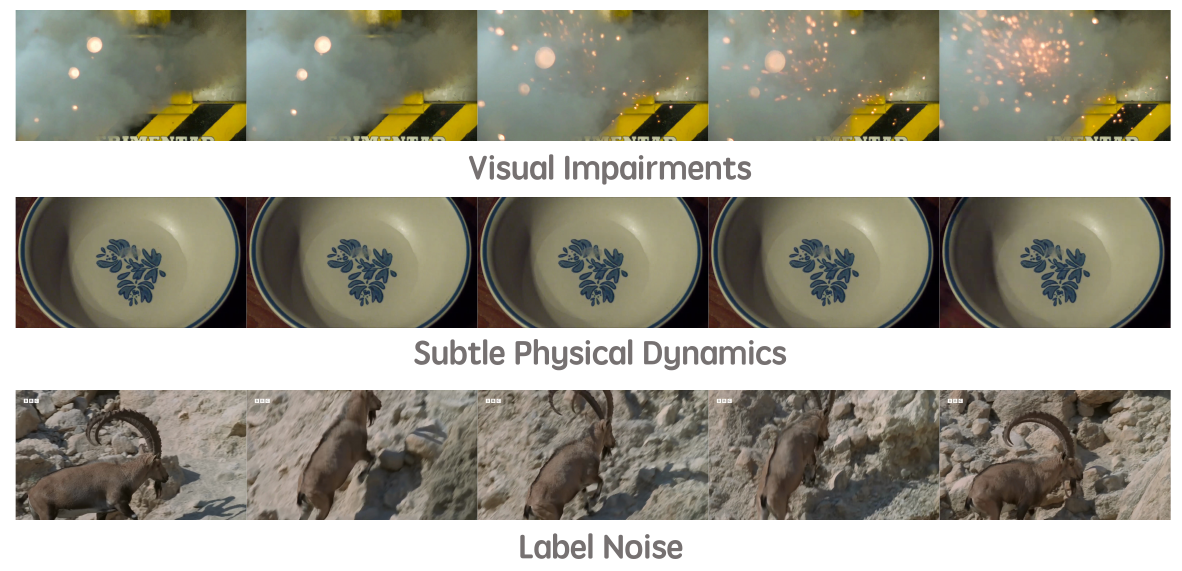}
    \caption{\textbf{Example of Video Artifacts} There are three main types of video artifacts: visual impairments, subtle physical dynamics and label noise.}
    \label{fig:sm_fig2}
\end{figure}

However, we observed that the raw video data in these subsets exhibit inconsistent quality. As illustrated in~\cref{fig:sm_fig2}, several common artifacts were identified during our preliminary check:
\begin{itemize}[label=\textbullet, leftmargin=*, nosep]
\item \textbf{Visual Impairments.} Significant motion blur or low resolution that hinders the learning of detailed dynamics.
\item \textbf{Subtle Physical Dynamics.} Instances where the physical phenomena are negligible or obscured by camera movement.
\item \textbf{Label Noise.} Cases of misclassification where the video content does not align with its assigned category.
\end{itemize}

To address the issue, we implemented a VLM-assisted data cleaning pipeline complemented by human verification. Specifically, we leveraged the multimodal understanding capabilities of Qwen2.5-VL-7B. For each data entry, the model was provided with the captions and three frames sampled per second from the video sequence. The model was then instructed to evaluate the data from 1 to 5, where 1 indicates very poor quality and 5 indicates excellent quality across three key dimensions:
\begin{itemize}[label=\textbullet, leftmargin=*, nosep]
\item \textbf{Text-Video Alignment.} Assessing the consistency between the video and the caption.
\item \textbf{Deformation Significance.} Evaluating the clarity and significance of the observed deformation phenomena.
\item \textbf{Visual Quality.} Identifying the presence of video artifacts such as black screen or severe motion blur.
\end{itemize}

\begin{figure}[h]
    \centering
    \includegraphics[width=1.0\linewidth]{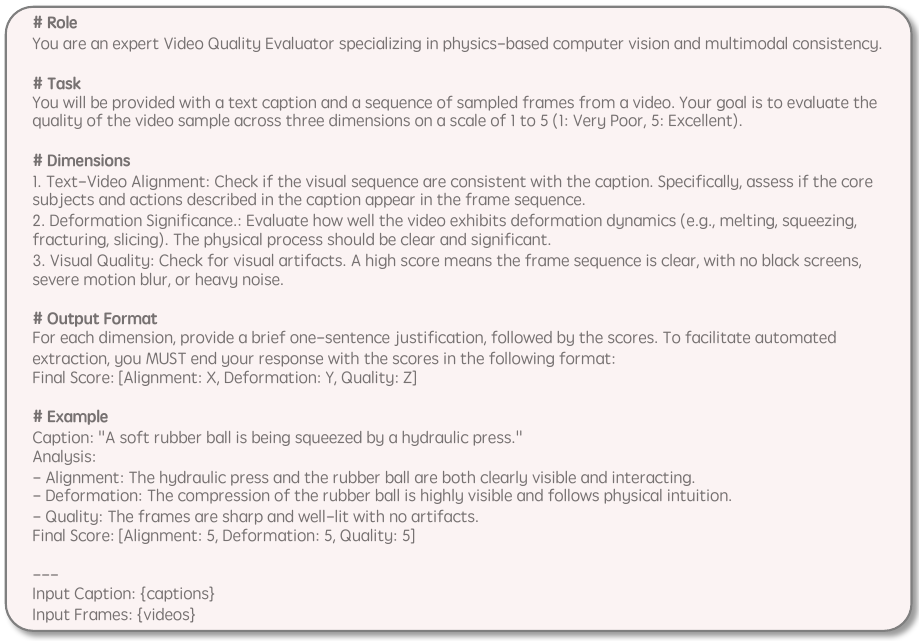}
    \caption{\textbf{Prompt Template for VLM-assisted Evaluation.} The detailed text prompt for Qwen2.5-VL-7B to evaluate data quality across three dimensions.}
    \label{fig:sm_fig3}
\end{figure}

The detailed prompt used for this evaluation is illustrated in~\cref{fig:sm_fig3}. For any sample scored 2 or lower in any of these dimensions, a manual review was conducted. If the human annotator confirmed the existence of the identified artifacts, the corresponding data sample was removed from the final dataset. 

\begin{figure}
    \centering
    \includegraphics[width=1.0\linewidth]{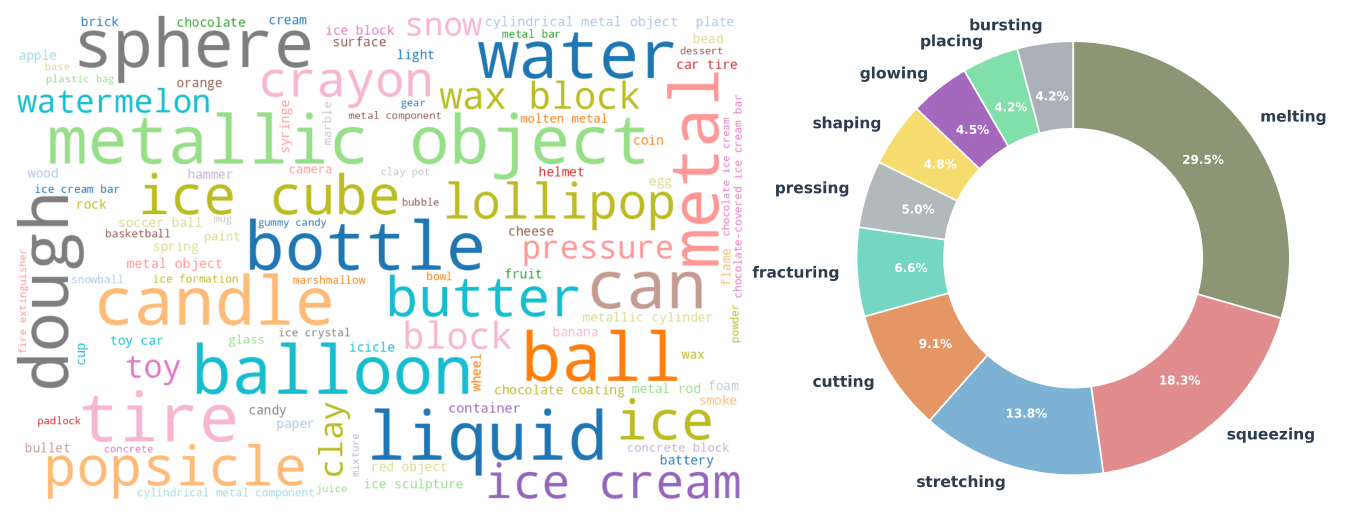}
    \caption{\textbf{Characteristics of Our Dataset.} The word cloud figure of core objects and the donut chart of dynamics verbs.}
    \label{fig:sm_fig4}
\end{figure}

We curated a total of 6,322 high-quality samples, with around 6k for training and 320 for evaluation. The characteristics of the dataset are visualized in the~\cref{fig:sm_fig4}. The word cloud figure illustrates the category distribution of core objects involved in physical phenomena, and a donut chart presents the distribution of dynamics verbs. These visualizations highlight the diversity of scenarios and deformation dynamics in our dataset.

\section{Details of Implementation}
\begin{figure}
    \centering
    \includegraphics[width=1.0\linewidth]{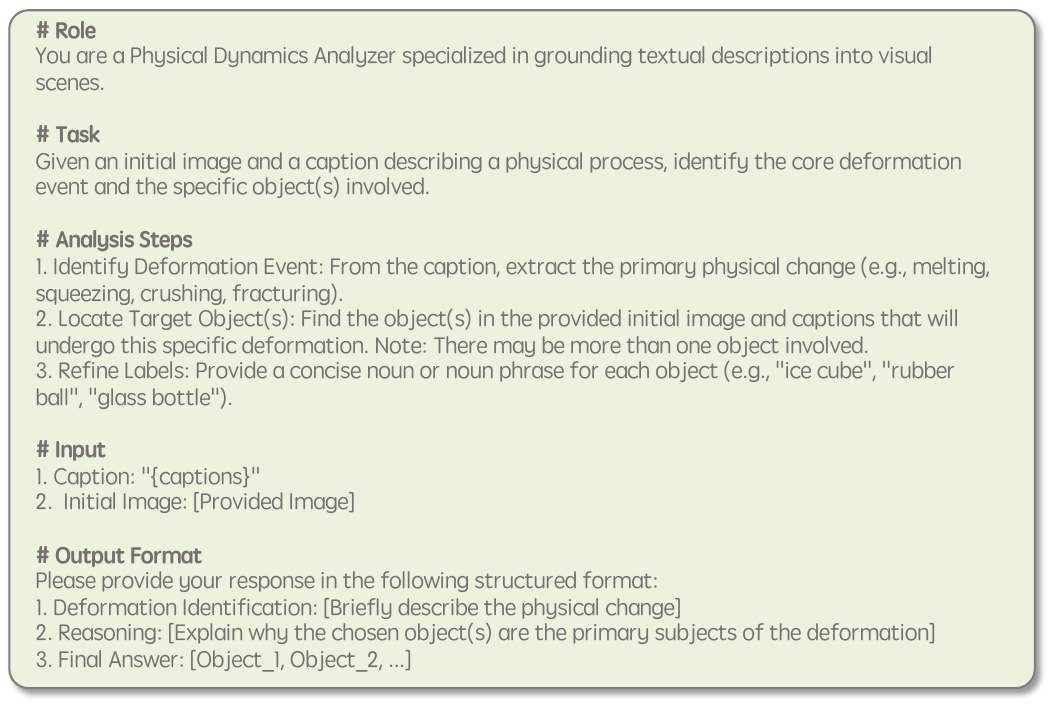}
    \caption{\textbf{Prompt Template.} The prompt template for identifying deformation objects.}
    \label{fig:sm_fig5}
\end{figure}

As discussed in~\cref{sec:DeforM-Reason}, we utilize Qwen2.5-VL-7B to identify the objects most likely to undergo deformation during both the training and inference stages. By providing the initial frame and the text caption, the model identifies the primary deformation objects using the prompt template illustrated in~\cref{fig:sm_fig5}. During the training stage, the identified object categories are processed with all video frames by the open-vocabulary object detector to ensure precise spatial localization. In the inference stage, the model similarly identifies the target object to retrieve its initial bounding box from the detector. After that, it plans and outputs the predicted trajectories for the bounding boxes in subsequent frames.

\section{Details of Evaluation}
\begin{figure}
    \centering
    \includegraphics[width=1.0\linewidth]{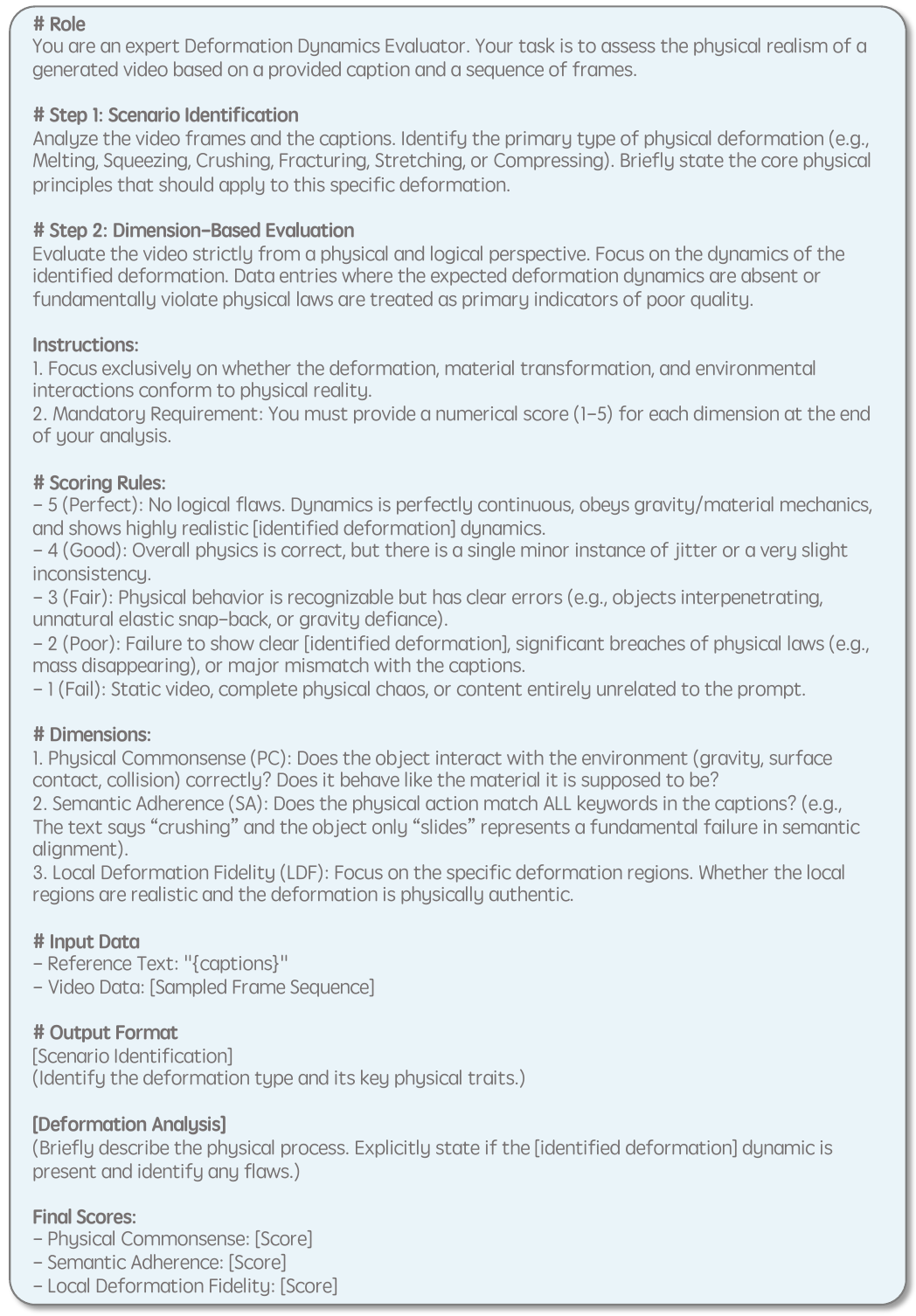}
    \caption{\textbf{Prompt Template.} The prompt template for physical consistency evaluation.}
    \label{fig:sm_fig6}
\end{figure}

As proposed in~\cref{sec:evaluation_settings}, we employ Qwen3-VL-32B to perform a comprehensive evaluation of the physical consistency of our generated results. For each test sample, the model is provided with the textual captions, the initial images, and the corresponding generated videos, which are sampled at a rate of 3 frames per second. We utilize the prompt template shown in~\cref{fig:sm_fig6} to quantitatively assess the model's performance across three key dimensions, Physical Commonsense (PC), Semantic Adherence (SA), and Local Deformation Fidelity (LDF).

\subsection{User Study}
To further evaluate the physical plausibility of our method, we conducted a user study focusing on \textit{Physical Consistency}. We invited 10 participants to perform a quantitative comparison between our method and four state-of-the-art baselines. As presented in~\cref{tab:user_study}, DeforM-Injection outperforms all baseline models by a significant margin across all evaluated dimensions, which demonstrates our model's superior capability in synthesizing complex deformation dynamics.

\begin{table*}[h]
\centering
\caption{User study of quantitative comparison on \textbf{Physical Consistency}. The yellow shaded column represents the aggregate Physical Score. The best results are highlighted in bold.}
\label{tab:user_study}
\resizebox{0.6\textwidth}{!}{
\begin{tabular}{l ccc >{\columncolor{hlyellow}}c}
\toprule
\rowcolor{white}
& \multicolumn{4}{c}{\textbf{Physical Consistency}} \\
\cmidrule(lr){2-5}
\rowcolor{white}
Method & PC $\uparrow$ & SA $\uparrow$ & LDF $\uparrow$ & \begin{tabular}[c]{@{}c@{}}Physical \\ Score $\uparrow$\end{tabular} \\
\midrule
CogVideoX1.5-I2V         & 2.86  & 3.36  & 3.00  & 3.07  \\
Hunyuan-I2V              & 2.88  & 2.32  & 2.72  & 2.64  \\
Wan2.2-I2V-14B           & 2.92  & 3.29  & 3.22  & 3.14  \\
MAGI-1                   & 2.68  & 2.40  & 2.56  & 2.54  \\
\midrule
\textbf{DeforM-Injection (Ours)} & \textbf{4.42}  & \textbf{4.38}  & \textbf{4.24}  & \textbf{4.34}  \\
\bottomrule
\end{tabular}
}
\end{table*}

\section{Use of Large Language Models}
We clarify the usage of large language models (LLMs) in the preparation of this work. LLMs were not used for research design or methodological designs. The use is limited to two aspects: (1) LLMs were utilized strictly to improve the linguistic clarity and grammatical accuracy. This process was limited to language polishing and did not involve the generation of new scientific ideas, research designs, or methodological frameworks.
(2) As illustrated in the main manuscript and supplemental material, VLMs act as integral parts of our framework and provide VLM-as-a-judge evaluation.

\end{document}